\documentclass{article}
\usepackage{ijcai21}
\usepackage{times}
\usepackage{soul}
\usepackage{url}
\usepackage[hidelinks]{hyperref}
\usepackage[utf8]{inputenc}
\usepackage[small]{caption}
\usepackage{graphicx}
\usepackage{amsmath}
\usepackage{amsthm}
\usepackage{booktabs}
\usepackage{algorithm}
\usepackage{algorithmic}
\usepackage{array}
\usepackage{adjustbox}
\usepackage{multirow}
\usepackage{multicol}
\usepackage{caption}
\usepackage{subcaption}

\title{An Examination of Fairness of AI Models for Deepfake Detection}

\author{
Loc Trinh\footnote{Contact Author} \hspace{0.8cm} Yan Liu\\
\affiliations
Department of Computer Science\\
University of Southern California\\
Los Angeles, CA 90089\\
\emails
\{loctrinh, yanliu.cs\}@usc.edu
}

\begin{document}

\maketitle
\begin{abstract}
Recent studies have demonstrated that deep learning models can discriminate based on protected classes like race and gender. In this work, we evaluate bias present in deepfake datasets and detection models across protected subgroups. Using facial datasets balanced by race and gender, we examine three popular deepfake detectors and find large disparities in predictive performances across races, with up to 10.7\% difference in error rate between subgroups. A closer look reveals that the widely used FaceForensics++ dataset is overwhelmingly composed of Caucasian subjects, with the majority being female Caucasians. Our investigation of the racial distribution of deepfakes reveals that the methods used to create deepfakes as positive training signals tend to produce ``irregular" faces - when a person’s face is swapped onto another person of a different race or gender. This causes detectors to learn spurious correlations between the foreground faces and fakeness. Moreover, when detectors are trained with the Blended Image (BI) dataset from Face X-Rays, we find that those detectors develop systematic discrimination towards certain racial subgroups, primarily female Asians.
\end{abstract}

\section{Introduction} \label{sec:intro}
Synthetic media have become so realistic with the advancement of deep neural networks that they are often indiscernible from authentic content. However, synthetic media designed to deceive poses a dangerous threat to many communities around the world \cite{Cahlan,Ingram}. In this context, \textit{Deepfake} videos – which portray human subjects with altered identities or malicious/embarrassing actions - have emerged as a vehicle for misinformation. With the current advancement and growing availability of computing resources, sophisticated deepfakes have become more pervasive, especially to generate revenge pornography \cite{Hao} and defame celebrities or political targets \cite{Vaccari}. Hence, there is a critical need for automated systems that can effectively combat misinformation on the internet.

To address this challenge, the vision community has conducted a series of excellent works on detecting deepfakes \cite{tolosana2020deepfakes,mirsky2021creation}. Sophisticated facial forgery detection tools \cite{Afchar,li2020face,liu2020global} and advanced training sets \cite{rossler2019faceforensics++v3,jiang2020deeperforensics10} were developed to train detectors capable of identifying deepfakes with high precision. Such results have also seen in real-world impact with Microsoft's release of Video Authenticator \cite{burt_horvitz_2020}, an automated tool trained on the publicly available FaceForensics++ dataset, to analyze a still photo or video to provide a percentage chance that the media is artificially manipulated. It works by detecting the blending boundary of the deepfake and subtle fading or grayscale elements that might not be detectable by the human eye. On the other hand, Facebook has also been pioneering its own system to detect AI-generated profiles and ban hundreds of fake accounts, pages, posts, and social groups\footnote{https://about.fb.com/news/2019/12/removing-coordinated-inauthentic-behavior-from-georgia-vietnam-and-the-us/}, along with strengthening its policy on deepfakes and authentic media\footnote{https://about.fb.com/news/2020/01/enforcing-against-manipulated-media/}.

While these works have achieved good progress towards the prediction task, detecting fake videos at a low false-positive rate is still a challenging problem~\cite{li2020face}. Moreover, since most studies focus on the visual artifacts existing within deepfakes, little is discussed about how such systems perform on diverse groups of real people across gender and race, which is the common setting where personal profiles and videos are being audited en masse for authenticity via automated systems. In this context, a small percentage difference in false-positive rates between subgroups would indicate that millions people of a particular group are more likely to be mistakenly classified as fake.

This draws a connection to fairness in machine learning, where growing concerns about unintended consequences from biased or flawed systems call for a careful and thorough examination of both datasets and models. Gender Shades \cite{gendershades} have demonstrated how facial recognition system discriminates across gender and race, showing a large gap in the accuracy of gender classifiers across different intersectional groups: darker-skinned females are misclassified in up to 34.7\% of cases, while the maximum error rate for lighter-skinned males is only 0.8\%. Others have shown that training with biased data has resulted in algorithmic discrimination \cite{bolukbasi2016man}. Although many works have studied how to create fairer algorithms, and benchmarked discrimination in various contexts \cite{Hardt2016EqualityOO,liu2018delayed}, few works have done this analysis for computer vision in the context of synthetic media and deepfakes. Our contributions are as follows:

\begin{enumerate}
    \item We find that the FaceForensics++ dataset commonly used for training deepfake detectors is overwhelmingly composed of Caucasian subjects, with the majority (36.6\%) of videos features female Caucasian subjects.
    \item We also find that approaches to generate fake samples as \textit{positive} training signals tend to overwhelmingly produce ``irregular" deepfakes - when a person’s face is swapped onto another person of a different race or gender, which leads to detectors learning spurious correlations between foreground faces and \textit{fakeness}.
    \item Using facial datasets balanced by gender and race, we find that classifiers designed to detect deepfakes have large predictive disparities across racial groups, with up to 10.7\% difference in error rate.
    \item Lastly, we observe that when detectors are trained with the Blended Images (BI) from Face X-Rays \cite{li2020face}, we find that detectors develops systematic discrimination towards female Asian subjects.
\end{enumerate}

\section{Related Work}

\subsection{Deepfake detection}
Early deepfake forensic work focused on hand-crafted facial features such as eye colors, light reflections, and 3D head poses / movements. However, these approaches do not scale well to more advanced GAN-based deepfakes. To combat the new generation of deepfakes, researchers leverage deep learning and convolutional networks to automatically extract meaningful features for face forgery detection \cite{rossler2019faceforensics++v3}. Work on shallow networks such as MesoInception4 \cite{Afchar} and Patch-based CNN \cite{chai2020makes} are developed to focus on low and medium level manipulation artifacts. Deep networks, such as Xception \cite{rossler2019faceforensics++v3}, also demonstrated success by achieving state-of-the-arts via fine-tuning on ImageNet. Other lines of research examine resolution-inconsistent facial artifacts DSP-FWA \cite{Li} through spatial pyramid pooling modules, blending artifacts via Face X-ray \cite{li2020face}, or temporal artifacts via dynamic prototypes \cite{trinhinterpretable}. FakeSpotter \cite{wang2020fakespotter} uses layer-wise neuron behaviors as features in addition to the output of the final-layer.

\subsection{Generalizability and robustness of detectors}
With more advanced deepfake creations, recent works \cite{cozzolino2018forensictransfer,8553251} have shown that the performance of current detection models \textit{drops} drastically on new types of facial manipulations. Few work call for a closer investigation into the generalizability of deepfake detectors towards unseen manipulations. In particular, ForensicTransfer \cite{cozzolino2018forensictransfer} proposes an autoencoder-based network to transfer knowledge between different but related manipulations via the hidden latent space. Face X-ray \cite{li2020face} addressed the problem by focusing on the more general blending artifacts as well as creating a blended image  dataset to help networks generalize across unseen manipulations. In addition to generalization, recent work have also demonstrated the vulnerability of deepfake detectors to adversarial attacks \cite{Carlini_2020_CVPR_Workshops}, where small tailored perturbations generated via either black-box or white-box attacks can easily fool the networks. This raises a concern about the robustness and commercial readiness of deepfake detectors. In contrast to complex adversarial attacks, our work examines the performances of deepfake detectors on natural images composing of different gender and diverse racial groups, as well as investigating the real-world consequences if deepfake detectors are commercially adopted.

\subsection{Algorithmic fairness and consequences}
Concerns about malicious applications of AI and unintended consequences from flawed or biased systems have propelled many investigations in studying representational and algorithmic bias. Gender Shades \cite{gendershades} have demonstrated how facial recognition system discriminates across gender and race, especially for darker-skinned females. \cite{liu2018delayed} showed that common fairness criteria may in fact harm underrepresented or disadvantaged groups due to delayed outcomes. \cite{celis2019controlling} proposed a framework to combat echo chambers created by highly personalized recommendations on social media that reinforced people’s biases and opinions. In terms of standardized approaches for the field,  \cite{mitchell2019model} and  \cite{gebru2018datasheets} recommend the usage of model cards and datasheets to better document the intended usage of models and data. Although many works have studied how to create fairer algorithms and benchmarked discrimination in various contexts \cite{Hardt2016EqualityOO,liu2018delayed}, we conduct a fairness analysis in the context of deepfake detection, which requires bookkeeping of the racial distribution face swaps and providing subgroup-specific deepfakes for audit.

\section{Deepfake Detection}
We investigate 3 popular deepfake detection models of various sizes, architectures, and loss formulations, all with proven success in detecting deepfake videos. We trained MesoInception4 \cite{Afchar}, Xception \cite{rossler2019faceforensics++v3}, and Face X-Ray \cite{li2020face} on the FaceForensics++ dataset, which contains four variants of face swaps. For a fair comparison, we also cross-test the models' generalizability across datasets with unknown manipulations not seen in FaceForensics++, such as Google's DeepfakeDetection, Celeb-DF, and DeeperForensics-1.0. Our results matche state-of-the-art results, which we then used to audit for fairness. For more detailed information on training, testing, and cross evaluations, see Section \ref{appendix:detection_methods} and Table \ref{tab:gen} in the Appendix.

\vspace{-0.1cm}

\begin{figure*}[!htb]
\centering
\includegraphics[keepaspectratio=false,width=0.9\textwidth]{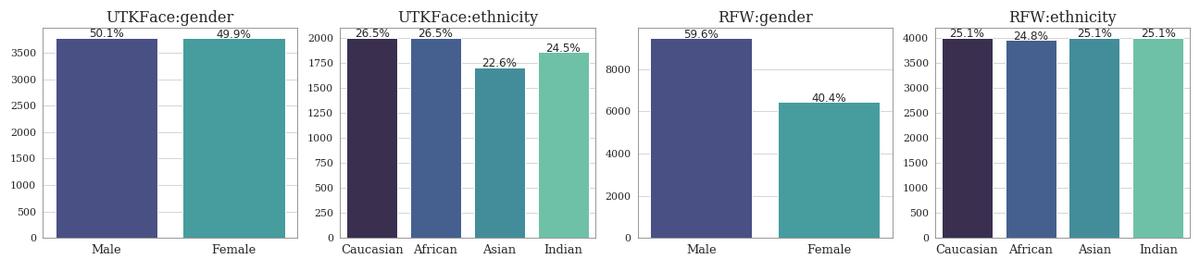}
\includegraphics[keepaspectratio=false,width=0.9\textwidth]{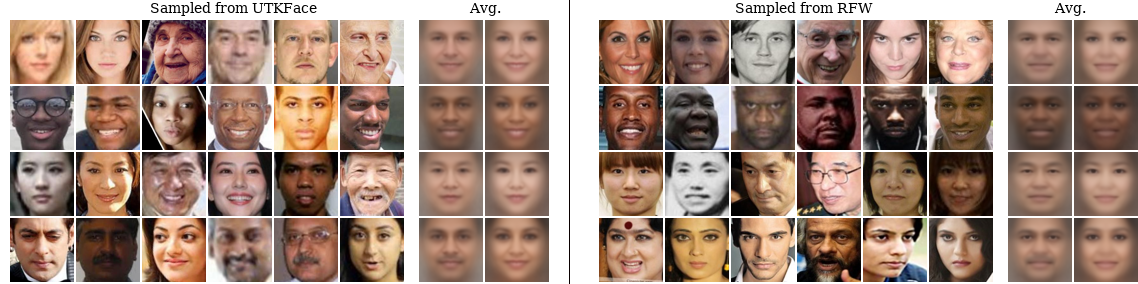}
\vspace{-0.2cm}
\caption{Examples and average faces of both the RFW and UTKFace database, along with their respective gender and racial distributions. In each row from top to bottom: Caucasian, African, Asian, Indian.}
\label{fig:rfwutk}
\vspace{-0.4cm}
\end{figure*}

\section{Deepfake Detection Audit}
We evaluated the deepfake detectors in the previous section, trained using both the FF++ and Blended Image (BI) datasets. Overall, all detectors perform equally on real and deepfake images containing male and female subjects, and all detectors trained with BI perform worst on media with darker African faces. Further analysis of the intersectional subgroups reveals that media with male African faces have the lowest TPR and media with female Asian faces have the highest FPR.

\subsection{Key findings on evaluated detectors}
\begin{itemize}
    \item All detectors perform equally on male faces and female faces (0.1 - 0.3\% difference in error rate)
    \item All detectors trained with BI perform worst on darker faces from the African subgroup, especially male African faces (3.5 - 6.7\% difference in error rate)
    \item For detectors trained with BI, faces from the Asian subgroup have the highest FPR, especially female Asian faces (5.2 - 8.1\% diff.)
    \item For detectors trained with BI, faces from the African subgroup have the lowest TPR, especially male African faces  (4.7 - 10.7\% diff.)
    \item FaceXRay + BI performs best on Caucasian faces, especially male Caucasian faces (9.8\%, 9.5\% error rate respectively). Meso4 and Xception detectors (with and without BI) perform best on Indian faces
    \item The maximum difference in error rate between the best and worst classified subgroups is 10.7\%
\end{itemize}

\subsection{Evaluation methodology}
We describe in detail the datasets and metrics utilized in this work to audit deepfake detectors. We adapted racially aware and fair facial recognition datasets labeled with demographic information for our task. For evaluations, we measure AUC and binary classification metrics across different subgroups.

\subsubsection{Auditing datasets}
We utilized two face datasets labeled with demographic information: (1) Racial Face-in-the-Wild (RFW) \cite{RFW} and (2) UTKFace \cite{UTKFace}. RFW is a dedicated testing dataset manually created for studying racial bias in face recognition. RFW contains four testing subsets, namely Caucasian, Asian, Indian, and African, with images selected from MS-Celeb-1M. Each subset contains about 10K images of 3K individuals for face verification - all with similar distribution with respect to age, gender, yaw pose, and pitch pose. Images in RFW have been carefully and manually cleaned. UTKFace is a large-scale face dataset with a long age span. The dataset consists of over 20K face images with annotations of age, gender, and race. The race labels consist of five groups, namely Caucasian, African, Asian, Indian, and Others (like Hispanic, Latino, Middle Eastern). All images in UTKFace cover large variations in pose, facial expression, illumination, occlusion, and resolution. 

For both datasets, we preprocessed images similarly to the deepfake images used for detection training (see Section \ref{appendix:preprocessing}).  Following RFW, we preserve the testing condition of the RFW dataset and do not alter the distribution of the images, which is adapted as the \textit{not-fake} portion of the testing dataset. For UTKFace, despite a large number of available images, it has a quite skewed distribution of racial groups. Hence, we did not use all labeled images but instead downsampled subgroups to achieve a balanced racial distribution similar to RFW. Figure \ref{fig:rfwutk} presents examples and average faces of both the RFW and UTKFace database, along with their respective gender and racial distributions.

\begin{figure}[!htb]
\vspace{-0.1cm}
\centering
\includegraphics[keepaspectratio=false,width=.9\columnwidth]{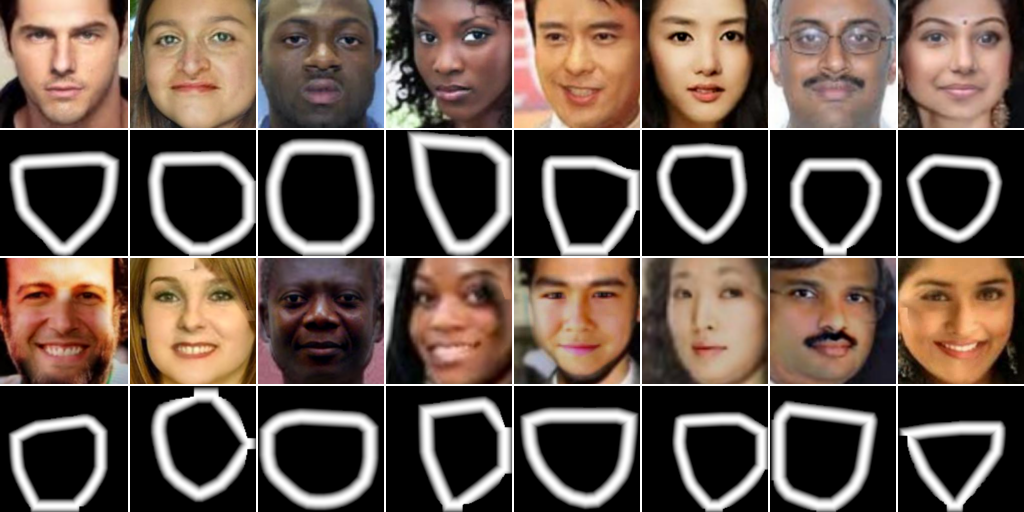}
\vspace{-0.2cm}
\caption{Visualized blended images along with their Face X-Rays for images with low (top row) and high (bottom row) artifacts.}
\label{fig:testing_samples}
\vspace{-0.5cm}
\end{figure}

To obtain \textit{deepfakes} for the testing dataset along with their demographic labels, we utilized the provided 68 facial landmarks within UTKFace to construct blended images, following the exact methodology as in \cite{li2020face}. To remain ethnically aware and also maintain demographic information, pairs of faces selected for swapping via the Face X-Rays approach are constrained to be from within the same gender and racial group. We generated 40K blended images per subgroup for a balanced distribution (Figure \ref{fig:testing_samples}). Our goal is to utilize a deepfake dataset with faithful demographic labels to audit the detectors' performance on both real and manipulated images.

\renewcommand{\arraystretch}{1.0}{
\begin{table*}[!htb]
\caption{Deepfake detection performance on gender and racial groups as measured by the area-under-the-ROC-curve (AUC), positive predictive value (PPV), error rate (1-PPV), true positive rate (TPR), and false positive rate (FPR) of the 3 evaluated deepfake detection models, trained using the standard and Blended Image (BI) approaches.}
\vspace{-0.1cm}
\label{tab:main}
\begin{adjustbox}{width=\textwidth}
\begin{tabular}{@{}ll|c|cc|cccc|cccccccc@{}} \toprule Model & Metric & ALL & M & F & Cau. & African & Asian & Indian & M/Cau. & F/Cau. & M/African & F/African & M/Asian & F/Asian & M/Indian & F/Indian \\ \midrule \multirow{5}{*}{Meso4} & AUC & 0.614 & \textbf{0.622} & 0.597 & 0.589 & 0.626 & 0.626 & 0.614 & 0.604 & 0.577 & 0.588 & 0.629 & \textbf{0.651} & 0.601 & 0.644 & 0.586 \\ & PPV & 0.784 & 0.764 & \textbf{0.802} & 0.779 & \textbf{0.794} & 0.789 & 0.774 & 0.795 & 0.764 & 0.661 & \textbf{0.940} & 0.807 & 0.772 & 0.797 & 0.754 \\ & Error Rate & 0.435 & \textbf{0.436} & 0.434 & \textbf{0.462} & 0.439 & 0.424 & 0.415 & 0.462 & \textbf{0.462} & 0.458 & 0.410 & 0.412 & 0.436 & 0.407 & 0.423 \\ & TPR & 0.541 & 0.517 & \textbf{0.565} & 0.493 & 0.524 & 0.555 & \textbf{0.592} & 0.476 & 0.510 & 0.457 & 0.591 & 0.556 & 0.554 & 0.578 & \textbf{0.605} \\ & FPR & 0.374 & 0.200 & \textbf{0.225} & 0.350 & 0.344 & 0.371 & \textbf{0.431} & 0.307 & 0.394 & 0.335 & 0.416 & 0.333 & 0.409 & 0.369 & \textbf{0.494} \\ \midrule \multirow{5}{*}{Xception} & AUC & 0.810 & 0.804 & \textbf{0.819} & 0.793 & 0.803 & 0.808 & \textbf{0.841} & 0.800 & 0.786 & 0.788 & 0.815 & 0.805 & 0.812 & 0.841 & \textbf{0.842} \\ & PPV & 0.856 & 0.827 & \textbf{0.888} & 0.863 & 0.838 & 0.861 & \textbf{0.865} & 0.864 & 0.861 & 0.739 & \textbf{0.957} & 0.857 & 0.865 & 0.863 & 0.868 \\ & Error Rate & 0.267 & \textbf{0.276} & 0.256 & \textbf{0.301} & 0.264 & 0.276 & 0.225 & 0.290 & \textbf{0.313} & 0.301 & 0.206 & 0.280 & 0.273 & 0.229 & 0.220 \\ & TPR & 0.753 & 0.749 & \textbf{0.758} & 0.687 & 0.783 & 0.731 & \textbf{0.812} & 0.704 & 0.671 & 0.755 & 0.812 & 0.730 & 0.733 & 0.808 & \textbf{0.816} \\ & FPR & 0.317 & \textbf{0.276} & 0.205 & 0.274 & \textbf{0.384} & 0.295 & 0.316 & 0.276 & 0.272 & 0.382 & \textbf{0.403} & 0.304 & 0.287 & 0.321 & 0.310 \\ \midrule \multirow{5}{*}{Meso4 + BI} & AUC & 0.795 & \textbf{0.811} & 0.765 & 0.798 & 0.79 & 0.775 & \textbf{0.821} & 0.818 & 0.785 & 0.766 & 0.78 & 0.82 & 0.73 & \textbf{0.845} & 0.8 \\ & PPV & 0.901 & \textbf{0.906} & 0.897 & 0.908 & \textbf{0.915} & 0.878 & 0.905 & 0.935 & 0.888 & 0.849 & \textbf{0.976} & 0.913 & 0.846 & 0.923 & 0.889 \\ & Error Rate & 0.356 & 0.355 & \textbf{0.357} & 0.362 & \textbf{0.385} & 0.358 & 0.319 & 0.384 & 0.34 & 0.367 & \textbf{0.413} & 0.341 & 0.374 & 0.324 & 0.314 \\ & TPR & 0.564 & 0.532 & \textbf{0.596} & 0.548 & 0.511 & 0.58 & \textbf{0.618} & 0.497 & 0.6 & 0.458 & 0.563 & 0.577 & 0.582 & 0.596 & \textbf{0.64} \\ & FPR & 0.155 & 0.062 & \textbf{0.105} & 0.138 & 0.12 & \textbf{0.201} & 0.162 & 0.087 & 0.19 & 0.116 & 0.153 & 0.138 & \textbf{0.264} & 0.123 & 0.2 \\ \midrule \multirow{5}{*}{Xception + BI} & AUC & 0.962 & \textbf{0.964} & 0.959 & 0.969 & 0.951 & 0.959 & \textbf{0.972} & 0.972 & 0.968 & 0.938 & 0.958 & 0.97 & 0.948 & \textbf{0.977} & 0.968 \\ & PPV & 0.952 & \textbf{0.956} & 0.949 & \textbf{0.963} & 0.957 & 0.933 & 0.957 & 0.971 & 0.955 & 0.928 & \textbf{0.987} & 0.956 & 0.912 & 0.969 & 0.945 \\ & Error Rate & 0.099 & 0.098 & \textbf{0.099} & 0.092 & \textbf{0.119} & 0.102 & 0.082 & 0.092 & 0.091 & \textbf{0.13} & 0.102 & 0.088 & 0.116 & 0.076 & 0.089 \\ & TPR & 0.907 & 0.896 & \textbf{0.919} & 0.907 & 0.873 & 0.923 & \textbf{0.926} & 0.898 & 0.916 & 0.846 & 0.901 & 0.918 & 0.927 & 0.923 & \textbf{0.93} \\ & FPR & 0.114 & 0.077 & \textbf{0.14} & 0.088 & 0.099 & \textbf{0.165} & 0.104 & 0.068 & 0.109 & 0.094 & 0.133 & 0.105 & \textbf{0.224} & 0.074 & 0.135 \\ \midrule \multirow{5}{*}{FaceXRay + BI} & AUC & 0.950 & 0.95 & 0.95 & \textbf{0.962} & 0.936 & 0.953 & 0.95 & \textbf{0.963} & 0.96 & 0.928 & 0.937 & 0.959 & 0.946 & 0.946 & 0.955 \\ & PPV & 0.939 & 0.932 & \textbf{0.947} & \textbf{0.951} & 0.944 & 0.931 & 0.933 & 0.95 & 0.952 & 0.906 & \textbf{0.985} & 0.938 & 0.923 & 0.932 & 0.933 \\ & Error Rate & 0.115 & \textbf{0.115} & 0.114 & 0.098 & \textbf{0.133} & 0.112 & 0.116 & 0.095 & 0.101 & \textbf{0.136} & 0.13 & 0.104 & 0.119 & 0.122 & 0.11 \\ & TPR & 0.897 & 0.895 & \textbf{0.899} & 0.91 & 0.865 & \textbf{0.912} & 0.902 & \textbf{0.915} & 0.905 & 0.858 & 0.872 & 0.914 & 0.909 & 0.894 & 0.911 \\ & FPR & 0.145 & 0.128 & \textbf{0.134} & 0.118 & 0.129 & \textbf{0.17} & 0.163 & 0.121 & 0.115 & 0.127 & 0.15 & 0.151 & \textbf{0.189} & 0.163 & 0.163 \\ \bottomrule \end{tabular} \end{adjustbox} \vspace{-0.3cm} \end{table*}}

\subsubsection{Evaluation metrics}
We analyze two sets of metrics, binary classification metrics and threshold agnostic Area under the ROC curve (AUC). For classification metrics, similar to Gender Shades \cite{gendershades}, we follow the gender classification evaluation precedent established by the National Institute for Standards and Technology (NIST) and assess the overall classification accuracy, along with the extension of true positive rate, false positive rate, and error rate (1-PPV) of the intersectional subgroups: \{male, female\} $\times$ \{Caucasian, African, Asian, Indian\}. Since the FaceForensics++ training dataset is heavily imbalanced, we set the threshold as the value in the range (0.01, 0.99, 0.01) that maximizes the balanced accuracy on the Faceforensics++ validation set. We also evaluated the AUC due to its robustness against class imbalance.
\vspace{-0.1cm}
\subsection{Audit results}
Table \ref{tab:main} shows detection performances on gender and racial groups as measured by the AUC, positive predictive value (PPV), error rate, true positive rate (TPR), and false positive rate (FPR) of the 3 deepfake detection models, trained using the FF++ and Blended Image (BI) datasets. We observe disparities in predictive performances between racial groups, which is most apparent in models trained with the BI dataset. 

\subsubsection{Gender groups audit}
From Table \ref{tab:main}, we observe that all detectors are \textit{equally} accurate in detecting manipulated images containing male and female subjects, with the difference in error rate as low as 0.1 - 0.3\%. For four out of five detectors, female subjects have both higher FPR and higher TPR. In the realistic setting where facial profiles on social media are automatically screened via deepfake detectors, FPR indicates that the proportion of real subjects mistakenly identified as fake can be much larger for female subjects than male subjects. This is especially true for the Xception + BI detector, which achieves the best result with error rates of 9.8\% on male subjects and 9.9\% on female subjects, but nearly twice as large FPR with 7.7\% for male subjects and 14.0\% for female subjects.

\begin{figure*}[!htb]
\centering
\includegraphics[keepaspectratio=false,width=0.95\textwidth]{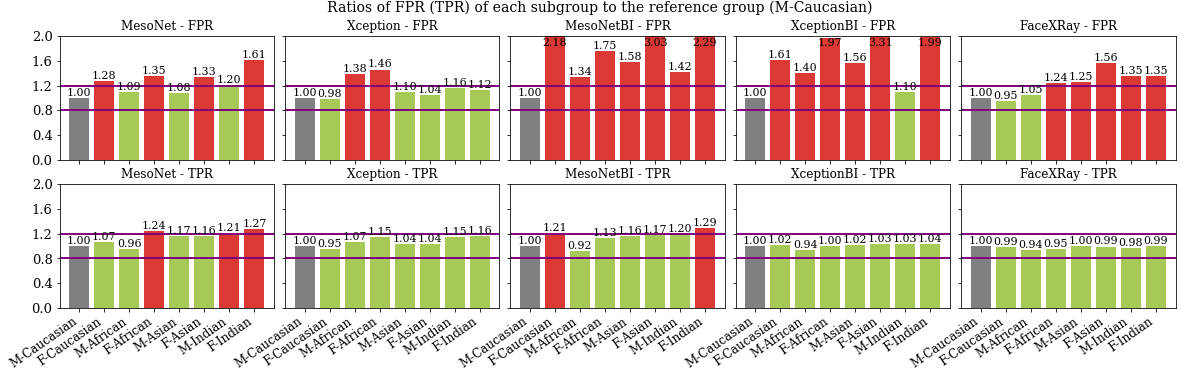}
\vspace{-0.2cm}
\caption{Ratios of FPR (TPR) for each of the intersectional subgroup to a reference group. In this case, we have chosen the ``M-Caucasian'' to be the reference group. Purple lines indicates the 20\% margins above and below. Red bars indicate the violation of these margins.}
\label{fig:TPRFPR}
\vspace{-0.3cm}
\end{figure*}

\subsubsection{Racial and intersectional subgroups audit}
We conduct an intersectional analysis of all detectors on all eight subgroups (M-Cau. F-Cau. M-African, F-African, M-Asian, F-Asian, M-Indian, and F-Indian). As seen in Table \ref{tab:main}, we observe large disparities in error rate across race, with the difference in error rate ranging from 3.5 - 7.6\% across all detectors. Of note, FaceXRay + BI performs best on Caucasian faces, especially male Caucasians (9.8\%, 9.5\% error rate respectively). MesoNet and Xception detectors (with and without BI) perform best on Indian faces. Across all detectors, the maximum difference in error rate between the best and worst intersectional subgroups is 10.7\%. Figure \ref{fig:TPRFPR} presents the ratios of FPR and TPR of each subgroup to a reference group, which we have chosen as the ``M-Caucasian'' group. We notice a stark contrast between FPR and TPR where we observe that subgroups with Asian or African racial backgrounds have false positive rates as high as three times as that of the reference group. In contrast, the TPRs of all groups are either well-within or around the accepted\footnote{Fairness for this metric is in [0.8, 1.2] w.r.t the Four-Fifths rule.} 20\% margins of the reference group, indicated by the purple lines.

In addition, there exists a trend in which all detectors trained with BI perform worst on African faces, especially male African faces with 3.5 - 6.7\% difference in error rate to the best subgroup. On a closer look, we can see that MesoNet + BI perform worst on female African faces, with 41.3\% error rate while Xception + BI and FaceXRay + BI perform worst on male African faces, with 13.0\% and 13.6\% error rate respectively. In addition to this trend, we also observe that, for detectors trained with BI, Asian faces have the highest FPR, especially female Asian faces with 5.2 - 8.1\% difference in FPR across all BI-trained detectors. Similarly, faces from the African subgroup have the lowest TPR, especially male African faces  4.7 - 10.7\% difference in TPR. This trend is uniquely consistent across all three detectors trained with BI (including the state-of-the-art Face X-Ray), even though the detectors all have diverse architectures and training losses.

\subsubsection{Analysis of results}
We agree with the findings in \cite{gendershades} that using single performance metrics such as AUC or detection accuracy over the entire dataset is not enough to justify massive commercial rollouts of deepfake detectors. Despite high AUC up to 0.962 and detection accuracy up to 90.1\% on our deepfake testing dataset, which would allow companies to claim commercial readiness for these detectors on all demographics represented, an intersectional analysis of the detectors shows otherwise. 

Our results also show indications of systematic bias in the learning process via generating and using manipulated images for training. Even though training with fake data generated via the BI process helped MesoNet, Xception, and Face X-Ray to improve their overall predictive performances, it also negatively impact predictions on real videos and images. Since fake artifacts are the focus of the detectors given how the training data was prepared, the absence of such artifacts in real and genuine media can lead to unintentional consequences in prediction. Figure \ref{fig:TPRFPR} plots the ratios of FPR (TPR) for each of the intersectional subgroup to a reference group. We have chosen the ``M-Caucasian'' to be the reference group. The disparities in FPR suggest that in a real-world scenario, facial profiles of female Asian or female African are 1.5-3 times more likely to be mistakenly labeled as fake than profiles of the male Caucasian. For large scale commercial applications, this would indicate bias against millions of people.

However, we note that the disparities observed are not ``intentionally'' built into the detectors. Figure \ref{fig:TPRFPR} (bottom) also demonstrated that the models are indeed focusing on manipulation artifacts as intended, where the the ratios of TPR across intersectional subgroups stay well within the 20\% margins around the reference group.  To the best of our knowledge, the closest work that mentions similar observations about performances between fake and real images is the work by Carlini and Farid \cite{Carlini_2020_CVPR_Workshops} where the authors use adversarial attacks to change the detectors' predictions. Noted as surprising by the authors, they found that it is harder to cause real images to be misclassified as fake, requiring up to 7\% of image pixels to be flipped, as opposed to causing fake images to be misclassified as real, needing just 1\% of pixels required to be flipped. We posit that because of the networks' focus on the detection of fake artifacts, it is easier to quickly fool the network using its gradient. However, the reverse direction is harder as the network has more trouble coming up with artifacts to ``manipulate" a real image.

\begin{figure*}[!htb]
\centering
\begin{subfigure}{.32\textwidth}
  \centering
  \includegraphics[width=\linewidth]{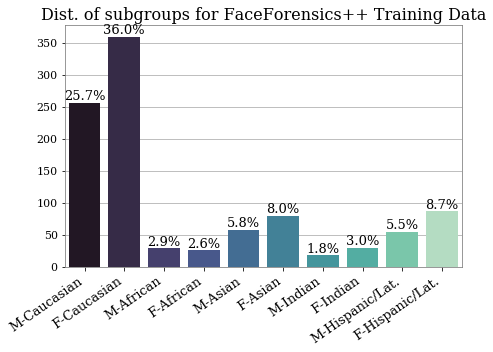}
  \label{fig:sub1}
\end{subfigure}%
\begin{subfigure}{.31\textwidth}
  \centering
  \includegraphics[width=\linewidth]{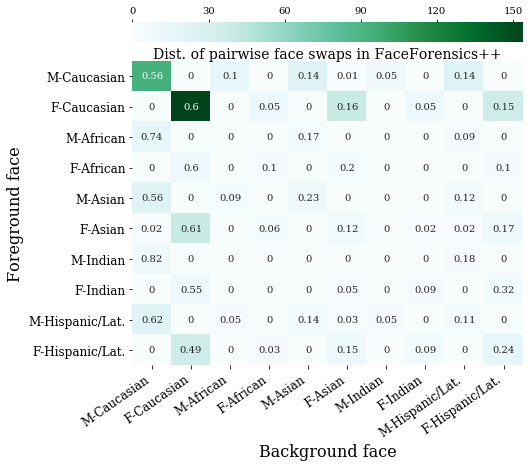}
  \label{fig:sub2}
\end{subfigure}%
\begin{subfigure}{.31\textwidth}
  \centering
  \includegraphics[width=\linewidth]{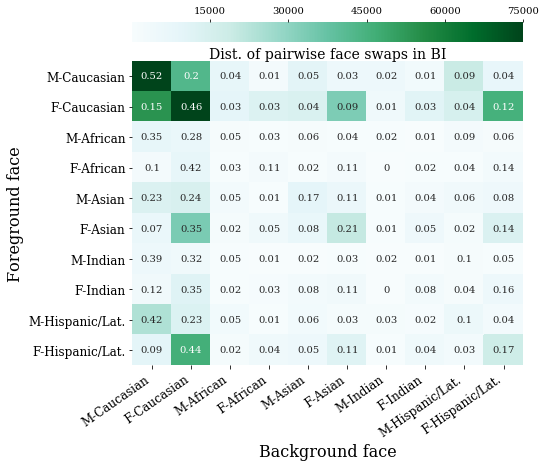}
  \label{fig:sub3}
\end{subfigure}
\vspace{-0.6cm}
\caption{(Left) Distribution of intersectional subgroups within FaceForensics++ real videos. (Middle, Right) Heatmaps of the distribution of pairwise swaps in FF++ and Blended Image (BI) datasets. Numbers in each row are normalized row-wise to present the percentage of swaps with foreground faces belonging to specific gender and racial group.}
\label{fig:training}
\vspace{-0.5cm}
\end{figure*}

\subsection{Training distribution and methodology bias}
To further investigate potential sources for bias in the trained detectors, we analyze both the FaceForensics++ and Blend Image (BI) datasets with respect to their gender and racial distribution. We observe the following key findings: 

\begin{itemize}
    \item Within FF++, 61.7\% of all real videos contain a person from the Caucasian group, with 36.0\% being female Caucasians.
    \item For FF++ fake videos, 59.44\% are videos of "irregular swaps". The rest are regular. "Irregular" swaps are when a person's face is swapped onto another person's face of a different gender or race.
    \item For BI blended face images, 65.45\% of the images are "irregular swaps". The rest are regular.
    \item For BI images with foreground female Asian faces, 35\% are swapped onto female Caucasian faces, 21\% onto female Asian faces, and 14\% onto female Hispanic faces.
\end{itemize}

\vspace{-0.2cm}
\subsubsection{Evaluation Methodology}
With the lack of demographic information for videos within the FaceForensics++ dataset, we manually collect ground truth demographic labels. To do so, we annotate each subject into two groups of perceived gender \{male, female\}, and five groups of perceived race \{Caucasian, African, Asian, Indian, Others\}. Three graduate annotators are selected for the task, with the assumption that each is of the same skill level in determining gender and racial group. For each subject, the annotators are presented with 5 distinct frames at various times in the video, which displayed the subject at different light angles and poses. We utilized pairwise percent agreement for multiple raters to measure Inter-Rater Reliability (IRR), and the majority label for each subject is selected as the ground truth demographic label. Our annotators achieve 75.93\% IRR, which is high for 2 genders and 5 racial groups.

With the demographic labels, we evaluate the percentage of ``regular'' and ``irregular'' faceswaps, where ``irregular'' is defined as a swap where a person's face is swapped onto another person's face of a different gender or race. FaceForensics++ provided the IDs for pairs of swaps for all four manipulation methods. Blended Images requires bookkeeping of the target and source faces selected via the BI methodology, which selects a source face from 5000 images where its 68 facial landmarks are closest in Euclidean distance to the target face.

\vspace{-0.1cm}
\subsubsection{Results}
Figure \ref{fig:training} presents the labeled distribution of intersectional subgroups within FaceForensics++ real videos. Overall, we observe a strong representation imbalance of gender and racial groups, with the videos containing 58.3\% of female subjects and 41.7\% of male subjects. The majority of authentic videos are of subjects from the Caucasian Group (61.7\%), with a major part being female Caucasians (36.0\%). Moreover, less than 5\% of the real videos contain subjects from the African group or Indian group, with the male Indian subgroup having the least representation. 

Figure \ref{fig:training} plots the heatmap of the distribution of pairwise swaps for manipulated videos/images in FF++ and Blended Image (BI) datasets. Numbers in each square are normalized row-wise to show the percentage of swaps with foreground faces belonging to specific gender and racial group. For FF++ fake videos, 59.44\% (428/720) are videos of ``irregular swaps", the rest are ``regular'' swaps (292/720). 58.75\% (423/720) of fakes have female Caucasians and male Caucasians subjects as foreground faces. Zooming in, we can see that the majority (60\%) (154/255 videos) of fakes with female Caucasian foreground faces are swapped onto other female Caucasians. On the other hand, the majority (61\%) (40/66 videos) of fakes with female Asian foreground faces are swapped onto female Caucasian faces, with only 7\% swapping onto female Asian faces. We also observed other types of irregular swaps where female faces are swapped onto male background faces. In the BI dataset, networks are trained with millions of swapped images. Here we sampled 1,000,000 images from the same process to visualize the distribution. Similarly, 65.45\% (654,400) of the images are ``irregular swaps", with the scale of BI is much more massive compared to FF++.  For BI blended face images with female Asian faces as foreground faces, the majority 35.3\% (34,031/96,443 images) are swapped onto female Caucasian faces, 21\% onto regular female Asian faces, and 14\% onto female Hispanic faces. Hence, given that the networks sees deepfakes with female Asian faces irregularly swapped for most of the time, it is more likely for them to learn a correlation between fakeness and Asian facial features. 

Without a specific way to pinpoint the exact source of bias, BI alone may not be fully responsible for misclassification and large disparities in false positive rates. However, we caution against using it for improving deepfake detection performances. Perhaps a more racially aware method to generate blended images could be essential for future directions.

\vspace{-0.1cm}
\section{Conclusion}

As deepfakes become more pervasive, there is a growing reliance on automated systems to combat deepfakes. We argue that practitioners should investigate all societal aspects and consequences of these high impact systems. In this work, we thoroughly measured the predictive performance of popular deepfake detectors on racially aware datasets balanced by gender and race. We found large disparities in predictive performances across races, as well as large representation bias in widely used FaceForensics++. Moreover, a majority of fakes are composed of ``irregular'' swaps between faces of different gender and races. Our work echoes the importance of benchmark representation and intersectional auditing for increased demographic transparency and accountability in AI systems.

\noindent \section*{Acknowledgment}
\noindent \small This work is supported by the Defense Advanced Research Projects Agency (DARPA) under Agreement No. HR00111990059. The authors would like to express their appreciation to colleagues Caroline Johnston and Nathan Dennler for many useful inputs and valuable comments on this work.

\small \bibliographystyle{named}
\bibliography{ijcai21}

\appendix

\section{Appendix}

{\renewcommand{\arraystretch}{1.1}
\begin{table}[ht!]
\centering
\caption{Basic information of training and testing datasets.}
\begin{adjustbox}{width=\columnwidth}
\begin{tabular}{@{}lcccc@{}}
\toprule
\multirow{2}{*}{Dataset} & \multicolumn{2}{c}{Real} & \multicolumn{2}{c}{Fake} \\  \cmidrule(lr){2-3}\cmidrule(lr){4-5}
& Video        & Frame        & Video         & Frame \\\midrule
FaceForensics++  (FF++) \cite{rossler2019faceforensics++v3} & 1000 & 509.9k & 4000 & 2039.6k \\
DeepFakeDetection (DFD) \cite{DDD_GoogleJigSaw2019} & 363 & 315.4k & 3068 & 2242.7k \\
DeeperForensics-1.0 \cite{jiang2020deeperforensics10} & 0 & 0.0k & 11000 & 5608.9k \\
Celeb-DF \cite{Li2019CelebDFAL} & 590 & 225.4k & 5639 & 2116.8k \\
\bottomrule
\end{tabular}
\end{adjustbox}
\label{tab:dataset}
\end{table}}

\renewcommand{\arraystretch}{1.1}{
\begin{table*}[ht!]
\caption{Test results on FaceForensics++ and cross-testing results on DeepFakeDetection (DFD), DeeperForensics, and Celeb-DF.}
\vspace{-0.3cm}
\label{tab:gen}
\begin{center}
\begin{adjustbox}{width=.85\textwidth}
\begin{tabular}{@{}lcccccccccccccc@{}} \toprule
Model & & \multicolumn{3}{c}{Test Dataset} & & \multicolumn{9}{c}{Cross-test Datasets} \\ \midrule
& & \multicolumn{3}{c}{FaceForensics++} & & \multicolumn{3}{c}{DFD} & \multicolumn{3}{c}{DeeperForensics} & \multicolumn{3}{c}{Celeb-DF} \\ \cmidrule(lr){3-5} \cmidrule(lr){7-9}\cmidrule(lr){10-12}\cmidrule(lr){13-15}
& & AUC & pAUC & ERR & & AUC & pAUC & ERR & AUC & pAUC & ERR & AUC & pAUC & ERR \\ \midrule
Meso4 & & 0.953 & 0.858 & 0.120 & & 0.828 & 0.766 & 0.258 & 0.822 & 0.742 & 0.269  & 0.729  & 0.575 & 0.339  \\
Xception & & 0.935 & 0.895 & 0.145 & & 0.913 & 0.813 & 0.166 & 0.941 & 0.818 & 0.124  & 0.778 & 0.564 & 0.291  \\ \midrule
Meso4 + BI & & 0.935  & 0.852 & 0.145 & & 0.883 & 0.779 & 0.204 & 0.881  & 0.803 & 0.184 & 0.713 & 0.559 & 0.329  \\
Xception + BI & & 0.989  & 0.985 & 0.029 & & 0.940 & 0.806 & 0.143 & 0.956  & 0.878 & 0.105 & 0.841 & 0.663 & 0.235  \\
FaceXRay + BI & & 0.992 & 0.989 & 0.023 & & 0.932 & 0.839 & 0.148 & 0.946 & 0.849 & 0.122 & 0.798 & 0.679 & 0.286\\ \bottomrule
\end{tabular}
\end{adjustbox}
\end{center}
\vspace{-0.3cm}
\end{table*}}

\subsection{Detection methods}
\label{appendix:detection_methods}
\paragraph{MesoInception4.} We first consider the class of shallow networks originally proposed to detect deepfakes via low-level features. Two commonly known approaches are MesoInception4 \cite{Afchar} and Patch-based CNN \cite{chai2020makes}. MesoInception4 is composed of very few convolutional layers, designed to focus on the mesoscopic properties of the images, as deeper networks tend to output more abstract and high-level representations. Patch-based CNN also uses shallow networks to learn local properties, except with the addition of 1v1 convolutions to accumulate patch-wise signals. Both have demonstrated success in predicting Deepfakes and variants such as NeuralTextures. Here, we chose the former for ease of implementation and code availability.

\paragraph{XceptionNet.} XceptionNet \cite{rossler2019faceforensics++v3} is one of the original deep benchmarks that performed well across four variants of face-swap deepfakes (Deepfakes, FaceSwap, Face2Face, and NeuralTextures). As the previous state-of-the-art, XceptionNet is a CNN trained on ImageNet, using separable convolutions with residual connections. The network is then fine-tuned on the FaceForensics++ dataset for the task of deepfake detection. Special preprocessing is required which involves detecting facial landmarks and pre-crop frames before feeding the images through XceptionNet.

\paragraph{Face X-Ray.} Face X-Ray \cite{li2020face} is the current state of the art model which leverages an advanced CNN backbone, High-Resolution Net (HRNet) and a novel segmentation approach to detect blending artifacts. In this work, additional fake images are generated procedurally by specifying a masked region to blend two faces. The procedure generates millions of deepfakes saved for training as additional data, termed Blended Images (BI) dataset. The known mask signals are then converted to Face X-Rays which contain the ground truth labeling of the swapping boundaries, which are then used as supervised training signals. 

Differing from the previous two models, which train only with the binary cross-entropy loss, Face X-Ray is additionally trained with the cross-entropy loss over Face X-Rays, which constrains the networks to produce soft binary map that focuses on the visual boundary artifacts.

\subsection{Training and testing datasets}
\label{appendix:training_testing}
For training the models, we use the FaceForensics++ (FF++) \cite{rossler2019faceforensics++v3} dataset, which consisted of 1000 real videos and 4000 fake videos that have been manipulated with four face manipulation methods: DeepFakes (DF)\footnote{https://github.com/deepfakes/faceswap}, Face2Face (F2F) \cite{face2face}, FaceSwap (FS)\footnote{https://github.com/MarekKowalski/FaceSwap/}, and NeuralTextures (NT) \cite{thies2019deferred}. FF++ also provides ground truth masks showing which part of the face was manipulated, which is useful for training Face X-Ray.

For testing evaluation, in addition to FF++, we also evaluate the cross-dataset generalizability of the models, using the following datasets: 1) DeepfakeDetection (DFD) \cite{DDD_GoogleJigSaw2019} which includes 363 real and 3068 deepfake videos released by Google in order to support developing deepfake detection methods; 2) DeeperForensics-1.0 \cite{jiang2020deeperforensics10} - large scale dataset consisting of 11,000 deepfake videos generated with paid actors compsing of  high-quality recordings of various identities, poses, expressions, emotions, and lighting conditionss; 3) Celeb-DF \cite{Li2019CelebDFAL}, a challenging DeepFake dataset of celebrities with 408 real videos and 795 synthesized videos with specially reduced visual artifacts. (Table~\ref{tab:dataset}).

\subsection{Pre-processing}
\label{appendix:preprocessing}
For pre-processing, we preprocess the videos by extracting the frames and cropping out subjects' faces based on facial landmarks detected by MTCNN \cite{mtcnn}. This exact pre-processing step was used for all models. In addition, we also trained another variant of both MesoInception4 and XceptionNet using the additional BI dataset for a fair comparison with FaceXray.

\subsection{Metrics and generalization results}
We train the models on the FaceForensics++ (FF++), and additionally cross-test on Google's DeepfakeDetection, Celeb-DF, and DeeperForensics-1.0, (Section \ref{appendix:training_testing} contains detailed training and testing information). To evaluate generalizability, we measure the area under the receiver operating curve (AUC) and the performance of deepfake detectors at a low false-positive rate threshold, using the standardized partial AUC or pAUC (at 10\%FPR). We additionally inspect the Equal Error Rate of the models, similarly to \cite{li2020face}. 

Table \ref{tab:gen} presents the test results on both FaceForensics++ and cross-testing results on DFD, DeeperForensics, and Celeb-DF for all trained models. The table shows that we matched results from previously published work, most notable on the challenging Celeb-DF dataset ($>$0.8 AUC). FaceXRay performed very well on FaceForensics++ and also generalizes to unseen manipulations during cross-testing. We notice that training with the Blended Image (BI) dataset also substantially improve performance for other architectures such as MesoInception4 and XceptionNet. This is expected as the BI dataset contains much more diverse facial forgeries along with a large amount of data augmentation including random downsampling and JPEG compression. With such impressive generalization results, we proceed to audit their performances across gender and racial groups.

\end{document}